\documentclass{article}

    \PassOptionsToPackage{numbers, compress}{natbib}

    \usepackage[preprint]{neurips_2022}

\usepackage{amsmath,amssymb} 
\usepackage[utf8]{inputenc} 
\usepackage[T1]{fontenc}    
\usepackage{url}            
\usepackage{booktabs}       
\usepackage{amsfonts}       
\usepackage{nicefrac}       
\usepackage{microtype}      
\usepackage{xcolor}         

\usepackage{tikz}
\usepackage{comment}
\usepackage{color}
\usepackage{todonotes}
\usepackage{multirow}
\usepackage{booktabs}
\usepackage[colorlinks, linkcolor=blue, anchorcolor=red, citecolor=green]{hyperref}

\title{Eye-gaze-guided Vision Transformer for Rectifying Shortcut Learning}

\author{%
  Chong Ma \\
  Northwestern Polytechnical University \\
  \texttt{mc-npu@mail.nwpu.edu.cn} \\
  \And
  Lin Zhao \\
  University of Georgia \\
  \texttt{lin.zhao@uga.edu} \\
  \And
  Yuzhong Chen \\
  University of Electronic Science and Technology of China \\
  \texttt{chenyuzhong211@gmail.com} \\
  \And
  Lu Zhang \\
  University of Texas at Arlington \\
  \texttt{lu.zhang2@mavs.uta.edu} \\
    \And
  Zhenxiang Xiao \\
  University of Electronic Science and Technology of China \\
  \texttt{zhenxiang.up@gmail.com} \\
    \And
  Haixing Dai \\
  University of Georgia \\
  \texttt{hd54134@uga.edu} \\
    \And
  David Liu \\
  Athens Academy \\
  \texttt{david.weizhong.liu@gmail.com} \\
    \And
  Zihao Wu \\
  University of Georgia \\
  \texttt{zw63397@uga.edu} \\
      \And
  Zhengliang Liu \\
  University of Georgia \\
  \texttt{zl18864@uga.edu} \\
      \And
  Sheng Wang \\
  Shanghai Jiao Tong University \\
  \texttt{wsheng@sjtu.edu.cn} \\
      \And
  Jiaxing Gao \\
  Northwestern Polytechnical University \\
  \texttt{2020262504@mail.nwpu.edu.cn} \\
      \And
  Changhe Li \\
  Northwestern Polytechnical University \\
  \texttt{ChangheLi@mail.nwpu.edu.cn} \\
      \And
  Xi Jiang \\
  University of Electronic Science and Technology of China \\
  \texttt{xijiang@uestc.edu.cn} \\
      \And
  Tuo Zhang \\
  Northwestern Polytechnical University \\
  \texttt{tuozhang@nwpu.edu.cn} \\
      \And
  Qian Wang \\
  ShanghaiTech University \\
  \texttt{wangqian2@shanghaitech.edu.cn} \\
      \And
  Dinggang Shen \\
  ShanghaiTech University \\
  \texttt{dgshen@shanghaitech.edu.cn} \\
      \And
  Dajiang Zhu \\
  University of Texas at Arlington \\
  \texttt{dajiang.zhu@uta.edu} \\
      \And
  Tianming Liu \\
  University of Georgia \\
  \texttt{tianming.liu@gmail.com} \\
}

\begin{document}

\maketitle

\clearpage
\begin{abstract}

Learning harmful shortcuts such as spurious correlations and biases prevents deep neural networks from learning the meaningful and useful representations, thus jeopardizing the generalizability and interpretability of the learned representation. The situation becomes even more serious in medical imaging, where the clinical data (e.g., MR images with pathology) are limited and scarce while the reliability, generalizability and transparency of the learned model are highly required. To address this problem, we propose to infuse human experts' intelligence and domain knowledge into the training of deep neural networks. The core idea is that we infuse the visual attention information from expert radiologists to proactively guide the deep model to focus on regions with potential pathology and avoid being trapped in learning harmful shortcuts. To do so, we propose a novel eye-gaze-guided vision transformer (EG-ViT) for diagnosis with limited medical image data. We mask the input image patches that are out of the radiologists' interest and add an additional residual connection in the last encoder layer of EG-ViT to maintain the correlations of all patches. The experiments on two public datasets of INbreast and SIIM-ACR demonstrate our EG-ViT model can effectively learn/transfer experts' domain knowledge and achieve much better performance than baselines. Meanwhile, it successfully rectifies the harmful shortcut learning and significantly improves the EG-ViT model's interpretability. In general, EG-ViT takes the advantages of both human expert's prior knowledge and the power of deep neural networks. This work opens new avenues for advancing current artificial intelligence paradigms by infusing human intelligence.


\end{abstract}

\section{Introduction}

Deep neural networks have been widely used and achieved remarkable successes in many fields including natural language processing, computer vision, and medical imaging \cite{lecun2015deep}, among others. Recent studies suggest that deep neural networks may be prone to learning the shortcut knowledge \cite{geirhos2020shortcut} such as the spurious correlations between the background and objects in the image (e.g., cows usually stand on the grass land) rather than intended relevant features. For example, some empirical works revealed that background is a harmful shortcut which drastically affects the deep learning model's performance in a negative way \cite{luo2021rectifying,xiao2020noise}. The harmful shortcut knowledge, on the one hand, may not be able to generalize to new domains and tasks, and thus degenerates the performance in some scenarios such as few-shot learning (FSL). On the other hand, it jeopardizes the interpretability of the model and prevents humans from validating its underlying reasoning which is crucial in many applications, e.g., disease diagnosis in medical imaging. 

\begin{figure*}[htb]
\begin{center}
\includegraphics[width=1.0\linewidth]{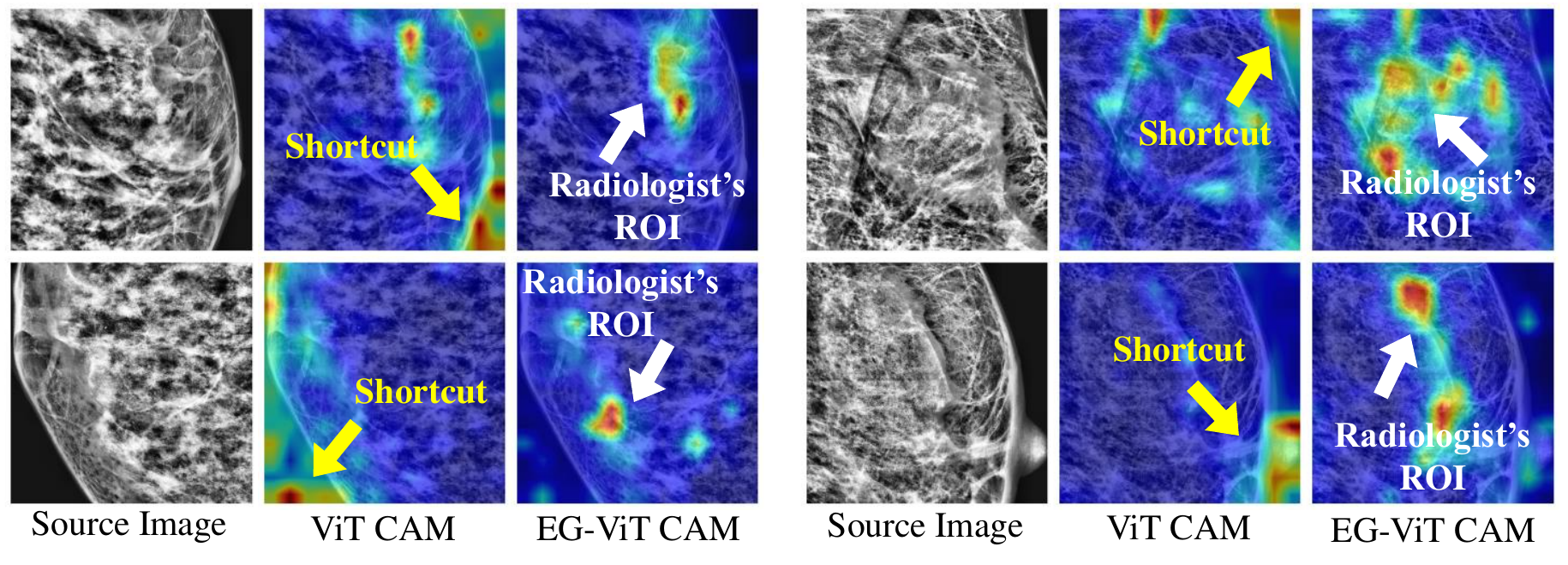}
\end{center}
\caption{Illustration of the shortcuts learned by ViT model. The left columns are the source images from the public INbreast dataset. The images in the middle column correspond to the model's attention derived from Grad-CAM. It is observed that the model focuses on background shortcuts (\textcolor{yellow}{yellow} arrows) rather than the valid breast tissues. The right columns are the Grad-CAM derived from our EG-ViT model. The regions of interests (ROIs) by radiologist are denoted by white arrows.}
\label{fig1}
\end{figure*}

Medical imaging analysis is a representative scenario where the harmful shortcut learning should be rectified because the generalizability and interpretability are highly desired and required, considering the scarcity of the clinical data (e.g., MR images with pathology) and the importance of reliability and transparency in clinical applications. The literature has already reported the existence of shortcuts in medical imaging applications \cite{luo2021rethinking,robinson2021deep,zech2018variable}. For example, in \cite{zech2018variable}, convolutional neural networks (CNNs) were employed to detect pneumonia and performed well with extremely high accuracy on the chest X-rays from a group of hospitals. However, it failed to generalize to the X-rays from other external hospitals with much lower performance: CNNs unexpectedly learned to detect a hospital-specific metal token at the corner of scans and utilized it for disease prediction indirectly \cite{zech2018variable,geirhos2020shortcut}. To motivate the work in this paper, in Fig.~\ref{fig1}, we also visualize four samples of harmful shortcuts learned by vision transformer (ViT) \cite{dosovitskiy2020image} model which are the medical images' background.

To solve this problem, one possible way is to enforce the model to concentrate on task-related objects or features by using prior knowledge \cite{luo2021rectifying}. For example, when conducting the diagnosis on medical images, the positions of eye-gaze from radiologists can be leveraged as additional domain knowledge. The rationale is that these positions are the regions-of-interest (ROIs) from a radiologist's perspective, which might be highly related to potential pathology. The domain knowledge embedded in ROIs from an expert is naturally interpretable and generalizable because it reaches the professional standard and has been validated and widely used in longstanding clinical practice. Some recent deep learning studies have already integrated eye-gaze of radiologists to improve the performance of medical image analysis \cite{2020Creation,wang2022follow}. However, an effective way of proactively infusing the expert's domain knowledge into deep learning processes to avoid harmful shortcut learning is still much needed.

In this paper, we propose an intuitive and effective method to infuse the domain knowledge of an expert with the training of deep learning models for FSL on disease diagnosis. Based on vision transformer (ViT) \cite{dosovitskiy2020image}, we introduce a novel eye-gaze-guided vision transformer (EG-ViT) model which applies an eye-gaze mask to input image patches to screen out those irrelevant to radiologist's visual attention and guide the model to focus on patches that are highly related to potential pathology. Meanwhile, a residual connection between the unmasked input and the last ViT encoder layer is intentionally added to retain the relationships of all patches. In this way, the EG-ViT model takes the advantages of both human expert's prior knowledge and the power of data-intensive ViT model, thus avoiding the harmful shortcut learning and infusing the expert's domain knowledge in a more effective manner. We evaluate the proposed EG-ViT on two public datasets, namely, INbreast \cite{InsMoreira2012INbreastTA} and SIIM-ACR~\cite{SIIM-ACR}. Our extensive experiments demonstrate that the proposed EG-ViT model significantly improves the diagnosis accuracy in the FSL scenarios and successfully avoids the harmful shortcut learning compared with the baseline ViT models (Fig.~\ref{fig1}).

In general, the main algorithmic and methodological novelties and contributions of our work are:

\begin{itemize}
\item We infuse the human expert's prior knowledge to guide the EG-ViT focusing on the patches with potential pathology. This design avoids the harmful shortcut learning and improves the generalizability and interpretability of the EG-ViT model with higher performance.

\item The proposed EG-ViT model only introduces the mask operation and an addition residual connection, thus allowing the inheritance of the parameters from a pre-trained vanilla ViT model without any additional cost.

\item Our method provides novel insights and a general framework for infusing human expert's domain knowledge into data-intensive and large-scale deep learning models such as ViT. Our work unlocks new paths for advancing current artificial intelligence paradigms by infusing human intelligence.

\end{itemize}

\section{Related Works}

\subsection{Shortcut Learning}
Deep neural networks often solve the task-specific problem, e.g., image classification, by learning the shortcuts such as the correlations of cows and grass instead of the intended solution, e.g., the features from cows \cite{geirhos2020shortcut}. Recently, the shortcut in deep learning models gains increasing attention across the deep learning field from computer vision (CV) \cite{dancette2021beyond,minderer2020automatic,xiao2020noise}, natural language processing (NLP) \cite{mccoy2019right,niven2019probing} to reinforcement learning \cite{amodei2016concrete}. To date, various methods have been devised to mitigate the negative effects of shortcuts \cite{luo2021rectifying}. For example, in \cite{luo2021rectifying}, a framework named COSOC was proposed to tackle this shortcut problem by extracting the foreground objects in images to get rid of background-related shortcuts based on a contrastive learning approach. \cite{du2021towards} proposed a measurement for quantifying the shortcut degree, with which a shortcut mitigation framework was introduced for natural language understanding (NLU). \cite{shen2021towards} forces the network to learn the necessary features for all the words in the input to alleviate the shortcut learning problem in supervised Paraphrase Identification (PI). In the medical imaging field, prior works also suggested the existence of shortcuts and proposed the strategies to neutralise shortcut learning such as removing the bias in the training dataset \cite{luo2021rethinking,nauta2022uncovering,robinson2021deep}.

\subsection{Vision Transformer}
Since ViT~\cite{dosovitskiy2020image} was published, transformer structure has been receiving increasing attention from the computer vision community~\cite{han2022survey}. Recently, several effective strategies have been proposed to improve model performance and efficiency in image classification, such as knowledge distillation in DeiT~\cite{touvron2021training}, depth-wise convolution in CeiT~\cite{yuan2021incorporating}, shifted windows in Swin Transformer~\cite{liu2021swin}, and tree-like structure in NesT~\cite{zhang2021aggregating}. However, the data-intensive characteristic of ViT makes it challenging to adapt to the target domain quickly with a limited amount of labeled data. The methods of distillation approach~\cite{touvron2021training}, smoothing the loss landscapes at convergence~\cite{chen2021vision}, and incorporating CNNs like CCT~\cite{hassani2021escaping} and local-ViT~\cite{li2021localvit} have been proposed to reduce the demand for extensive training data to a certain extent. Nonetheless, fast adaption to the target domain for FSL still requires more innovative and effective methods to reduce the demand for training data. 

In the medical imaging domain, ViT-style models have been explored in computer-aided diagnosis tasks on the chest X-ray (CXR) images~\cite{shamshad2022transformers}. Krishnan et al.~\cite{krishnan2021vision} and Park et al.~\cite{park2021vision} utilize ViT-based models to achieve higher COVID-19 classification accuracy through CXR images. COVID-Transformer~\cite{shome2021covid} and xViTCOS~\cite{mondal2021xvitcos} have been proposed to further improve classification accuracy and focus on diagnosis-related regions. However, there is still much room for improvement to train ViT models in a small dataset, such as medical imaging dataset.

\subsection{Eye Tracking in Radiology}
\label{eye tracking}

Visual diagnosis plays a central role in radiology, and eye-tracking procedures have proven to be a valuable tool in the study of visual diagnostic processes in radiology for decades~\cite{2010Current}. 
Back in 1978, Nodine et al.~\cite{CalvinFNodine1987UsingEM} proposed a three-stage theory of visual search and detection, that is, distinguishing between initial overall pattern recognition, focused attention to image details, and final decision making. Then, the global-focal search models~\cite{RichardGSwensson1980ATD,HaroldLKundel2007HolisticCO,TraftonDrew2013InformaticsIR} further optimized the interpretation of eye-movement behavior. They also found that experts can quickly locate potential lesions with a global search and use a larger functional field of view and more conceptual knowledge than novices to find abnormalities. 
Krupinski~\cite{ElizabethAKrupinski1996VisualSP} reported that in mammograms with more than one lesion, experts showed more comparison scanning between the left and right breast. 
Kok et al.~\cite{EllenMKok2012LookingIT} found that experts' visual search patterns were more diffuse than novices. 
Then the authors~\cite{EllenMKok2017BeforeYV} showed that experts searched normal CXR more systematically than novices. 
Overall, existing literature studies of eye movement in radiology and inter-individual variation provide sufficient justification to integrate this wealth of ancillary information in computer-aided diagnostic (CAD) systems in radiology.

With the rise of deep learning in CAD, the combination with eye movement is also increasing. 
Mall et al.~\cite{SuneetaMall2018ModelingVS} modeled the visual search behavior of radiologists and their interpretation of mammography using CNNs. 
Furthermore, they~\cite{SuneetaMall2019MissedCA} investigated the relationship between human visual attention and CNNs in ﬁnding lesions in mammography. 
Recently, Karargyris et al.~\cite{AlexandrosKarargyris2021CreationAV} developed a dataset with CXR, gaze, and text diagnosis reports. 
They proposed a multi-task framework which predicted gaze and diagnosed diseases at the same time.
Wang et al.~\cite{wang2022follow} used radiologists' visual attention to supervise the CNN's attention via an attention consistency module, thus improving the diagnosis performance in osteoarthritis assessment of knee X-ray images.

\subsection{Few-Shot Learning in Medical Imaging}
FSL is proposed for learning with only a few samples~\cite{wang2020generalizing}. Therefore, FSL is quite suitable for medical image classification due to the difficulty of obtaining a dataset with large and consistently labeled medical images. Ma et al.~\cite{ma2019affinitynet} adopts a k-Nearest-Neighbor attention pooling layer to construct the AffinityNet model, which can learn from a small number of training data and perform well in generalization. Puch et al.~\cite{puch2019few} applies an FSL model based on Triplet Network to accomplish classification of brain images. MetaCOVID~\cite{shorfuzzaman2021metacovid}, which is an FSL model based on Siamese Network, is proposed to classify COVID-19 cases from CXR images, and achieves a promising performance. The methods mentioned above mainly uses CNNs to extract features and they may have limited interpretability of the model. Li et al.~\cite{li2021few} present polymorphic transformer, which exists between feature extractor and a task head and reduces the gap between different domains. This work achieves high performance on medical segmentation tasks and shows the flexibility of transformer in FSL. Although transformer-based FSL has made great progress in computer vision, more works are needed to take the full advantage of the powerful representation learning capability of ViT in medical imaging.
\section{Method}

The method section is organized as follows: We firstly illustrate the design of EG-ViT in Section~\ref{model}. Then, we introduce the pre-processing of eye-gaze data and the generation of eye-gaze mask in Section \ref{gaze_mask}.

\subsection{Eye-Gaze-Guided Vision Transformer}
\label{model}

\begin{figure*}[htb]
\begin{center}
\includegraphics[width=1.0\linewidth]{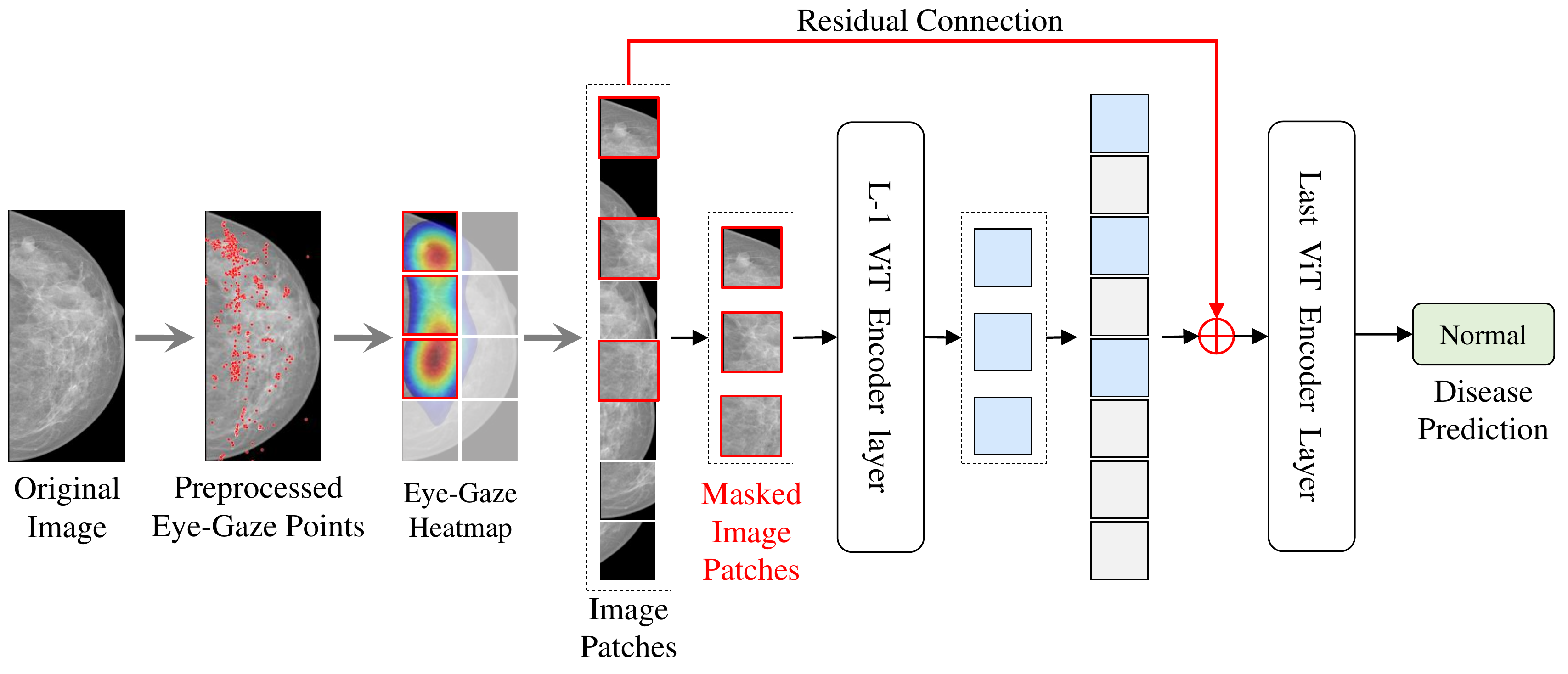}
\end{center}
\caption{The architecture of our EG-ViT model. The eye-gaze points are collected and pre-processed to generate the eye-gaze heat map as masks. Then, we applied the mask to the patches derived from the original image to screen out the patches that may not be related to pathology and radiologists' interest. The masked image patches (highlighted by \textcolor{red}{red} rectangular) are treated as input to the transformer encoder layer. Note that to maintain the relationships among all the patches, we added an additional residual connection (highlighted by \textcolor{red}{red} arrow) between the input and the last encoder layer.}
\label{EGMG-ViT}
\end{figure*}

We introduce the detailed architecture of the proposed EG-ViT in this subsection. Compared with natural images, medical images generally have a higher resolution while pathology such as lesions usually locates in a small region with a noisy background, making it difficult for the extraction of the meaningful features. To avoid learning the useless or harmful shortcuts rather than intended meaningful features, an intuitive idea is to guide the model to focus on the regions that are potentially related to pathology based on prior domain knowledge. The visual attention from a radiologist during the diagnosis can provide such domain knowledge as the guidance for model training. In this paper, we implement this idea by introducing a eye-gaze guided mask on input image patches and an additional residual connection on the EG-ViT model. The architecture of EG-ViT model is shown in Fig.~\ref{EGMG-ViT}.

\subsubsection{Eye-Gaze Guided Mask Operation}
With the eye-gaze guided mask, we can perform a mask operation on the input patches of the ViT model. Specifically, the input image can be divided into $N$ patches where $N=(H \times W)/P^2$ is the patch numbers, $H$ and $W$ are the height and weight of images, $P$ is the patch size. The ViT model maps the images patches $x_p^i$ ($i =1,2,\cdots,$N) to \emph{D} dimension patch embedding $z_0 \in \mathbb{R}^{(N+1) \times D}$ (contacted with a class token) with a trainable linear projection $E \in \mathbb{R}^{HWC \times D}$ where $C$ is the channels of the images:

\begin{equation}
\label{eq_1}
    z_0 = [x_{class};x_p^1E;x_p^2E;\cdots;x_p^NE]+E_{pos}
\end{equation}
where $z_0^0=x_{class} \in \mathbb{R}^N$ is the $class$ token for classification and $E_{pos} \in \mathbb{R}^{(N+1) \times D}$ is the learnable position embedding. Then, the embedding of the input image patch $z_0$ is masked as:
\begin{equation}
\label{mask patch}
    \Tilde{z}_0=[z_0^0;z_0^{1:N}\odot mask]
\end{equation}
where $mask \in \mathbb{R}^N$ is the binary eye gaze mask detailed in Section~\ref{gaze_mask} and $z_0^{1:N}=[x_p^1E;x_p^2E;\cdots;x_p^NE]$ is the image embedding patches. The masked patch embedding $\Tilde{z}_0$ is then input into the first layer of ViT encoder, forcing the model focus on the corresponding patches with potential pathology.

\subsubsection{Residual Connections Preserving Global Features}
For vision transformer \cite{dosovitskiy2020image}, the forward propagation of each transformer encoder layer can be written as:
\begin{align}
    \Tilde{z}_l^\prime &=MSA(LN(\Tilde{z}_{l-1}))+\Tilde{z}_{l-1}\\
    \Tilde{z}_l   &=MLP(LN(\Tilde{z}_l^\prime))+\Tilde{z}_l^\prime
\end{align}
where $\Tilde{z}_l^\prime$ is the \emph{l}-th layer's masked embedding patches. $MSA$, $MLP$ and $LN$ are the multiheaded self-attention, multilayer perceptron, and layer norm in each block.

However, masking some patches in the first layer results in a risk of missing useful background information and positional relationships among blocks. Inspired by \cite{He_2016_CVPR,he2021masked}, we add the whole initial embedding patches back to the last layer's embedding patches to retain global information and maintain the correlations of all patches. Therefore, the input embedding patch of last transformer encoder  $\hat{z}_{l-1}^i$ ($i$=0,1,2 $\cdots$ N) can be written as:
\begin{align}
    \hat{z}_{l-1}^i=\begin{cases} 
    \Tilde{z}_{l-1}^0,      &{if} \; i=0 \\ 
    z_0^i,          &{if} \; mask_i=0 \\ 
    \Tilde{z}_{l-1}^i+z_0^i,&otherwise\end{cases}
\end{align}
where $\hat{z}_{l-1}$ and $\Tilde{z}_{l-1}$ are the after and before additive operations of the embedding patches before the last transformer encoder.

\subsubsection{Pre-training and Fine-tuning Style}
It should be noted that we do not add any additional parameters to the model, thus allowing our model to inherit the parameters of the pre-trained model directly without additional operation. Also, the computation of the transformer model is related to the number of patches, and by adding a mask, we also reduce the computation of ViT. Pre-training models on large datasets and fine-tuning on specific dataset is a popular paradigm for FSL. In this study, we initialize the model parameters with the model pre-trained on ImageNet-1K~\cite{OlgaRussakovsky2014ImageNetLS}.

\subsection{Generation of Eye-Gaze Guided Mask}
\label{gaze_mask}
As discussed in Section~\ref{model}, the visual attention of radiologist plays a central role in our EG-ViT model which provides a wealth of expert's domain knowledge. Here, we introduce the steps to generate the eye-gaze guided mask used in our model.

\begin{figure*}[htb]
\begin{center}
\includegraphics[width=0.9\linewidth]{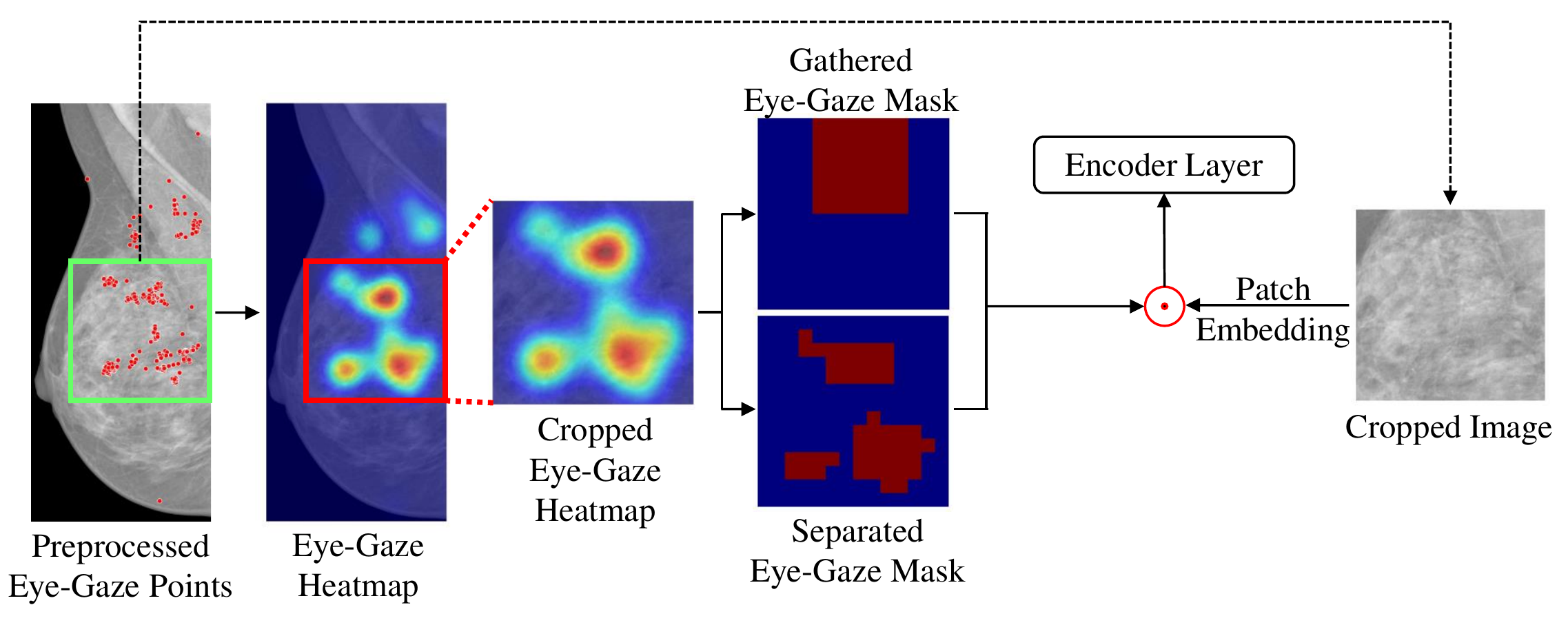}
\end{center}
\caption{The generation of gathered/separated eye-gaze masks. The eye-gaze heat map is firstly cropped into the size of 224$\times$224, based on which gathered/separated eye-gaze masks are generated, respectively. Then, these eye-gaze masks are used to mask the corresponding patch embedding, which is the input to the EG-ViT encoder layer.}
\label{fig3}
\end{figure*}

Firstly, the eye-tracking data (i.e., the eye-gaze points) is collected when the radiologist read the medical images such as X-ray for diagnosis. More details for eye-tracking data acquisition can be found in supplemental materials. With the raw eye gaze points from the eye tracker, we filter the eye movements such as Saccades, Smooth Pursuits Nystagmus \cite{robinson1968oculomotor}, and only keep Fixations of radiologists (red dots in Fig.~\ref{fig3}) by using the well-established I2MC \cite{nystrom2010adaptive} method.

Then, the eye-gaze heatmap is generated by Gaussian filtering of the eye-gaze scatter points as shown in Fig. \ref{fig3}. Next, to facilitate the training of the model, we crop a ROI from both the original image and eye-gaze heat maps. As the patch embedding layer maps the input image to $14 \times 14$, we also resize the eye-gaze heat map to $14 \times 14$.

Last, as shown in Fig. \ref{fig3}, the binary Separated Eye-Gaze Mask is obtained by setting the largest 49 values to 1 and zeroing the rest. The Gathered Eye-Gaze Mask is defined as a $7 \times 7$ binary mask centered at the largest value. The gathered mask can filter out the areas but the greatest interest to the radiologist while the separated mask tends to focus on all sub-regions of interest to the radiologist. 

Notably, the combination of eye-gaze guided mask generation and mask-guided vision transformer (EG-ViT) offers an effective and powerful mechanism to infuse radiologist's domain knowledge into the representation learning of the most informative, explainable and generalizable features for medical diagnosis. Our experimental results in the next section will demonstrate that this EG-ViT model can effectively rectify harmful shortcut learning and thus significantly improve the quality of learned features with relatively small training dataset.

\section{Experiments}

To demonstrate the value of introducing human experts' experience and prior knowledge into deep neural network training, we apply the proposed EG-ViT model to two different public clinical datasets (Section~\ref{Datasets}) with eye tracking data from human radiologists. We firstly compare the prediction performance with two baselines, ResNet~\cite{He_2016_CVPR} and Vision Transformer~\cite{dosovitskiy2020image} in Section~\ref{ComparisonBaselines}. Secondly, we provide a detailed analysis of how EG-ViT can rectify shortcut in Section~\ref{EvaluationShortcut}. Thirdly, we compare with other classification methods that used eye gaze data in Section~\ref{ComparisonGazeClassification}, followed by ablation study in Section~\ref{Ablation}.

\subsection{Datasets and Evaluation Metrics}
\label{Datasets}
We conduct the experiments on two datasets: INbreast \cite{InsMoreira2012INbreastTA} and SIIM-ACR \cite{KhaledSaab2021ObservationalSF,SIIM-ACR}. The INbreast dataset \cite{InsMoreira2012INbreastTA} includes 410 full-ﬁeld digital mammography images which were collected during low-dose X-ray irradiation of the breast. And we invited a radiologist with 10 years of experience to diagnose the images in this database and collected the complete eye movement data using our acquisition system. According to BI-RADS~\cite{LauraLiberman2002BreastIR} assessment of masses, the images can be classified into three groups: normal(302), benign(37) and malignant(71), respectively. Saab et al.~\cite{KhaledSaab2021ObservationalSF} randomly selected 1,170 images, with 268 cases of pneumon, from the SIIM-ACR Pneumothorax dataset~\cite{SIIM-ACR} and collected gaze data from three radiologists. We randomly split the dataset into 80\% and 20\% as training and testing dataset. For each experiment, we report the accuracy (ACC), F1-score (F1) and area under curve (AUC) on test dataset.

\subsection{Implementation Details}

We train 70 epochs with 0.0001 initial learning rate and choose a cosine decay learning rate scheduler and 8 epochs of warm-up. An Adam optimizer with a batch size of 64 are applied in our study. 
For the INbreast dataset \cite{InsMoreira2012INbreastTA}, images and corresponding eye-gaze heatmap are with resolution of $3000 \times 4000$ pixels. For a better performance, we crop the ROI images as input images. For images with masses (include benign and malignant classes), we only crop the ROI containing the mass area as a sample. Since there are multiple masses within a single image, we count the two immediately adjacent masses as just one ROI, and finally obtain a total of 114 ROIs of mass in 108 images, of which 40 are benign and 74 are malignant. Considering the larger image field of view, we choose the size of ROI as $1024\times1024$. For the normal category without mass, we randomly crop only one ROI image for each image. Therefore, we obtain 302 normal, 40 benign and 74 malignant samples. The INbreast dataset \cite{InsMoreira2012INbreastTA} is divided into training and testing dataset based on samples instead of images. To balance the numbers of training samples of each category, we randomly crop the ROI 8 times for benign category and 4 times for malignant category, that is, the mass will appear at random location in the ROI images. Note that, for benign and malignant case in testing sets, the mass is centered in ROI image. Finally, the training dataset contains a total of 482 normal samples, 512 benign mass samples, and 472 malignant mass samples. We also use \cite{CLAHE} to perform contrast enhancement twice with different threshold parameters for each cropped image in training sets. The cropped ROI images are both resized to $224 \times 224$ pixels finally.
For the SIIM-ACR dataset~\cite{SIIM-ACR}, the eye-gaze heat map and X-ray images are simply resized to $224 \times 224$ pixels with no ROI cropping.

\subsection{Comparison with Baselines}
\label{ComparisonBaselines}
Here, we adopt ResNet \cite{He_2016_CVPR} and Vision Transformer \cite{dosovitskiy2020image} as two baselines for comparison. The baseline model was pre-trained on ImageNet dataset \cite{deng2009imagenet} and fine-tuned on the two clinical dataset. The experimental results and parameters of each model are reported in Table \ref{table1}. On the INbreast dataset \cite{InsMoreira2012INbreastTA}, our EG-ViT model uses a relatively small number of parameters. Our model outperforms all the baselines in terms of F1 scores. We also compare with two baseline ViT models, ViT-S and ViT-B. Among them, ViT-S use 384 as hidden size and 12 encoder layers, ViT-B use 768 as hidden size and 12 encoder layers. It is observed that the performance of the two baseline ViT models is inferior to that of ResNets. This is because ViT has a more complex model architecture as well as a larger function space than ResNet and lacks some inductive biases that only CNNs have, making it difficult to pre-train and fine-tune on a small dataset like INbreast \cite{InsMoreira2012INbreastTA}. However, with the guidance of eye-gaze from the radiologist, the performance of ViT-based EG-ViT model is significantly improved, which suggests that eye-gaze serves as a strong prior guidance to assist network training and reduces the potential over-fitting problem caused by insufficient samples.

\setlength{\tabcolsep}{4pt}
\begin{table}
\begin{center}
\caption{The disease prediction accuracy compared with baseline ResNet and ViT model in terms of Accuracy, F1 and AUC scores. The number of parameters in each model is also reported. \textcolor{red}{Red} and \textcolor{blue}{blue} denote the best and the second-best results, respectively.}
\label{table1}
\begin{tabular}{cccccccc}
\hline\noalign{\smallskip}
\multirow{2}{*}{Method}&\multirow{2}{*}{Params}&\multicolumn{3}{c}{INbreast \cite{InsMoreira2012INbreastTA}} & \multicolumn{3}{c}{SIIM-ACR~\cite{SIIM-ACR}}\\ \cmidrule(r){3-5} \cmidrule(r){6-8}
  & & Acc. & F1 & AUC  & Acc. & F1 & AUC \\
\noalign{\smallskip}
\hline
\noalign{\smallskip}
ResNet-18~\cite{He_2016_CVPR}         &11M & 84.34 & 85.55 & 89.04   & 71.67 & 67.85 & 72.42\\ 
ResNet-50~\cite{He_2016_CVPR}         &24M & 91.57 & 91.92 & \textcolor{red}{94.62}   & 69.17 & 67.3 & 69.25\\ 
ResNet-101~\cite{He_2016_CVPR}        &43M & \textcolor{blue}{92.07} & \textcolor{blue}{92.55} & 91.52   & 70.83 & 65.33 & 70.31\\
ViT-S~\cite{dosovitskiy2020image}             &22M & 83.13 & 81.06 & 85.36   & 81.20 & \textcolor{blue}{81.62} & \textcolor{blue}{74.46}\\
ViT-B~\cite{dosovitskiy2020image}            &89M & 87.95 & 81.35 &90.86   & \textcolor{red}{86.00} & 73.69 & \textcolor{red}{86.67}\\
EG-ViT (ours)         &22M & \textcolor{red}{92.77} & \textcolor{red}{92.81} & \textcolor{blue}{94.16}   & \textcolor{blue}{85.60} & \textcolor{red}{84.87} & 74.10\\

\hline
\end{tabular}
\end{center}
\end{table}
\setlength{\tabcolsep}{1.4pt}

On the SIIM-ACR dataset~\cite{SIIM-ACR}, EG-ViT outperforms all the other methods in F1 measure, and ViT-B has higher accuracy and AUC score than the rest of the models. However, ViT-B has a much larger number of parameters and is more difficult to train, while EG-ViT is smaller and achieves essentially the same level of accuracy as the larger model.

\subsection{Evaluation of Shortcut Rectification}
\label{EvaluationShortcut}

\begin{figure*}
\begin{center}
\includegraphics[width=0.9\linewidth]{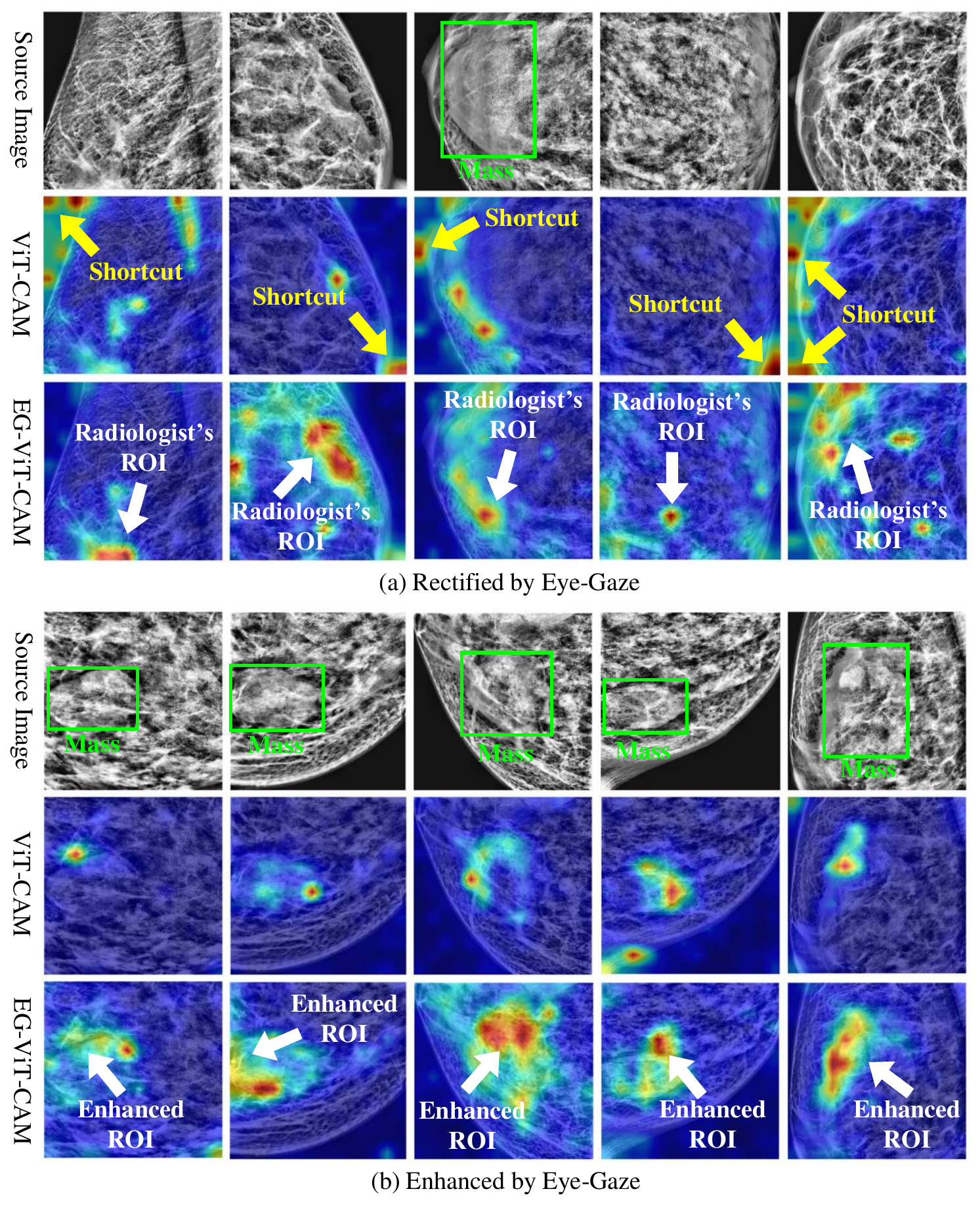}
\end{center}
\caption{(a) Harmful shortcut learning rectified by eye gaze guidance. (b) Useful feature learning enhanced by eye gaze guidance. In each panel of (a) and (b), the first row is the source image, the second row is the attention map of ViT obtained using Grad-CAM, and the third row is the attention map of EG-ViT. Each column corresponds to the same example.}
\label{fig_rectify}
\end{figure*}

In this subsection, we evaluate the performance of EG-ViT model in rectifying the shortcut learning both qualitatively and quantitatively. For better visualization of the comparison, we employ the Grad-CAM module \cite{Selvaraju_2017_ICCV} to generate the network attention map. Grad-CAM uses gradient to calculate the attention map of the model, which does not require any changes to the model structure and thus can be easily deployed to the ViT model.

Fig. \ref{fig_rectify} shows two ways of rectification by our EG-ViT model for qualitative comparison. In Fig. \ref{fig_rectify}(a), the source images are shown in the first row. The Grad-CAM maps from fine-tuned vanilla ViT model and from EG-ViT are demonstrated in the second and third rows, respectively. It is observed that without the expert's domain knowledge, ViT model is likely to make classification decisions from the areas that are related to regions, such as background edge, other than valid human tissues. However, with the help of visual attention from radiologists, the EG-ViT model is guided to the meaningful areas, such as the inner mammary region. Fig. \ref{fig_rectify}(b) demonstrates the cases that EG-ViT model guides and enhances the model's attention to the radiologist's ROI. The regions with mass are more emphasized by EG-ViT compared with vanilla ViT, which makes the decision of the model more interpretable.
To measure the improvement of EG-ViT model quantitatively, we manually check the Grad-CAM to examine if the harmful shortcuts exist in ViT and are rectified by EG-ViT in INbreast dataset \cite{InsMoreira2012INbreastTA}. We find that EG-ViT has a significant improvement on 64\% (264 cases) of all 410 cases compared with ViT. Among them, 151 cases are significantly corrected and 113 are significantly enhanced. Therefore, the radiologist's attention plays a crucial role in shortcut rectification. 

\subsection{Comparison with Eye-Gazed based Classification Network}
\label{ComparisonGazeClassification}

\setlength{\tabcolsep}{4pt}
\begin{table}
\begin{center}
\caption{Comparison with other eye-gaze-guided networks. \textcolor{red}{Red} and \textcolor{blue}{blue} denote the best and the second-best results, respectively.}
\label{table2}
\begin{tabular}{ccccccc}
\hline\noalign{\smallskip}
\multirow{2}{*}{Method}&\multicolumn{3}{c}{INbreast \cite{InsMoreira2012INbreastTA}} & \multicolumn{3}{c}{SIIM-ACR~\cite{SIIM-ACR}}\\ \cmidrule(r){2-4} \cmidrule(r){5-7}
& Acc. & F1 & AUC  & Acc. & F1 & AUC \\
\noalign{\smallskip}
\hline
\noalign{\smallskip}
ResNet-18+Gaze \cite{wang2022follow}          & 86.74 & 88.32 & 91.48   & 70.0 & 70.07 & 68.94\\
ResNet-50+Gaze \cite{wang2022follow}         & 90.36 & 90.95 & 93.06   & 70.83 & 69.22 & 70.58\\
ResNet-101+Gaze \cite{wang2022follow}         & \textcolor{blue}{91.57} & \textcolor{blue}{91.92} & \textcolor{red}{94.62}   & 72.5 & 71.22 & \textcolor{blue}{72.66}\\
U-Net+Gaze \cite{2020Creation}              & 86.07 & 85.36 & 92.98   & \textcolor{blue}{81.10} & \textcolor{blue}{80.33} & 68.84\\
EG-ViT (ours)                 & \textcolor{red}{92.77} & \textcolor{red}{92.81} & \textcolor{blue}{94.16}   & \textcolor{red}{85.60} & \textcolor{red}{84.87} & \textcolor{red}{74.10}\\
\hline
\end{tabular}
\end{center}
\end{table}
\setlength{\tabcolsep}{1.4pt}

In this subsection, we compare our EG-ViT model with two recent studies that also utilized eye-gaze for medical image classification tasks \cite{wang2022follow,2020Creation} . In \cite{wang2022follow},  ResNet \cite{He_2016_CVPR} was used as the classification backbone, with which visual attention from eye-gaze data was incorporated to enhance osteoarthritis assessment on knee X-ray images. Karargyris et al. \cite{2020Creation} used a U-Net structure to classify three chest diseases. And the model also can output attention map to compare with human attention map. We train these two methods for the disease classification task on INbreast \cite{InsMoreira2012INbreastTA} and SIIM-ACR \cite{SIIM-ACR}, respectively. As shown in Table~\ref{table2}, our proposed EG-ViT model outperforms the other methods on both INbreast \cite{InsMoreira2012INbreastTA} and SIIM-ACR datasets~\cite{SIIM-ACR} in terms of most metrics, except for AUC for INbreast \cite{InsMoreira2012INbreastTA}, on which our model yields the second-best performance. It can be seen that the aid of eye-gaze heat map can improve the performance of ViT model on small datasets, even beyond the CNN models.

\subsection{Ablation Study}
\label{Ablation}

\setlength{\tabcolsep}{4pt}
\begin{table}
\begin{center}
\caption{Comparison of performance using different masks. \textcolor{red}{Red} and \textcolor{blue}{blue} denote the best and the second-best results, respectively.}
\label{table3}
\begin{tabular}{ccccccc}
\hline\noalign{\smallskip}
\multirow{2}{*}{Method}&\multicolumn{3}{c}{INbreast \cite{InsMoreira2012INbreastTA}} & \multicolumn{3}{c}{SIIM-ACR~\cite{SIIM-ACR}}\\ \cmidrule(r){2-4} \cmidrule(r){5-7} & Acc. & F1 & AUC  & Acc. & F1 & AUC \\
\noalign{\smallskip}
\hline
\noalign{\smallskip}
ViT-S~\cite{dosovitskiy2020image} (Baseline)  
&83.13 &81.06 &85.36   &81.20 &\textcolor{blue}{81.62} &\textcolor{red}{74.46}  \\
Gathered Grad-CAM Mask
&83.13 &84.08 &\textcolor{blue}{90.36}   &82.00 &76.54 &57.71\\
Separated Grad-CAM Mask
& 83.13 & \textcolor{blue}{84.13} & 89.65  & 82.40 & 79.74 & 64.18\\
Gathered Eye Gaze Mask
& 79.52 & 81.45 & 88.89   & \textcolor{blue}{84.00} & 81.58 & 65.9\\
Separated Eye Gaze Mask
& \textcolor{red}{92.77} & \textcolor{red}{92.81} & \textcolor{red}{94.16}   & \textcolor{red}{85.6} & \textcolor{red}{84.87} & \textcolor{blue}{74.1}\\
\hline
\end{tabular}
\end{center}
\end{table}
\setlength{\tabcolsep}{1.4pt}

In this subsection, we discuss the effect of the eye gaze mask. As shown in Fig.~\ref{fig3}, we conduct comparative experiments on separated position mask and gathered position mask. The comparison of performance using these two types of masks is shown in Table \ref{table3}. The first row of the table shows the results of training with the ViT-S~\cite{dosovitskiy2020image}. In the second and third rows, we use the Grad-CAM generated by the network and compare the effect of gathered and separated position masks, and the fourth and fifth rows show the comparison of the two masks generated by human attention heat-map. In our experiments on both datasets, we find that the separated Grad-CAM mask is slightly better than the gathered one, especially for eye gaze mask. This could be attributed to that separated regions have the advantage of helping the model learn the relationship between features that are further apart in larger images. Also, when a radiologist has strong prior knowledge in the region out of the mask, the eye gaze mask will have a more scattered distribution, due to the flexibility of human attention itself and the different reading patterns of experts. We also see that the result of gathered gaze position masks is worse than Grad-CAM mask on the INbreast dataset \cite{InsMoreira2012INbreastTA}. This may be related to the radiologists' individualized reading habits. Specifically, if the radiologists' gaze points are spread out and the saccade path is long, then the use of a gathered position mask will ignore the features at the other locations where the radiologists focus on. We will continue to optimize the way that gaze mask is generated when we collect more eye gaze data from radiologists in the future.

\section{Conclusion}

In this paper, we proposed a novel eye-gaze-guided vision transformer (EG-ViT) to infuse human expert’s intelligence and domain knowledge into the training of deep neural networks. This EG-ViT model is designed and implemented via the combination of eye-gaze guided mask generation and mask-guided vision transformer. The experiments on the INbreast \cite{InsMoreira2012INbreastTA} and SIIM-ACR~\cite{SIIM-ACR} datasets demonstrated that radiologist's visual attention can effectively guide the EG-ViT model to concentrate on regions with potential pathology and achieve much better performance compared with baselines. In particular, our EG-ViT model successfully rectifies the harmful shortcut learning and significantly improves the model's interpretability, generalizability, and performance. Overall, this work contributes novel insights and a concrete new method for advancing current artificial intelligence paradigms by infusing human intelligence. Our future works include extending and evaluating the EG-ViT framework on other types of images, e.g., natural images, with eye-tracking data for few-shot learning problems and various downstream tasks.

%
%
\bibliographystyle{splncs04}
\bibliography{references}

\begin{thebibliography}{10}
\providecommand{\url}[1]{\texttt{#1}}
\providecommand{\urlprefix}{URL }
\providecommand{\doi}[1]{https://doi.org/#1}

\bibitem{amodei2016concrete}
Amodei, D., Olah, C., Steinhardt, J., Christiano, P., Schulman, J., Man{\'e},
  D.: Concrete problems in ai safety. arXiv preprint arXiv:1606.06565  (2016)

\bibitem{chen2021vision}
Chen, X., Hsieh, C.J., Gong, B.: When vision transformers outperform resnets
  without pretraining or strong data augmentations. arXiv preprint
  arXiv:2106.01548  (2021)

\bibitem{dancette2021beyond}
Dancette, C., Cadene, R., Teney, D., Cord, M.: Beyond question-based biases:
  Assessing multimodal shortcut learning in visual question answering. In:
  Proceedings of the IEEE/CVF International Conference on Computer Vision. pp.
  1574--1583 (2021)

\bibitem{deng2009imagenet}
Deng, J., Dong, W., Socher, R., Li, L.J., Li, K., Fei-Fei, L.: Imagenet: A
  large-scale hierarchical image database. In: 2009 IEEE conference on computer
  vision and pattern recognition. pp. 248--255. Ieee (2009)

\bibitem{dosovitskiy2020image}
Dosovitskiy, A., Beyer, L., Kolesnikov, A., Weissenborn, D., Zhai, X.,
  Unterthiner, T., Dehghani, M., Minderer, M., Heigold, G., Gelly, S., et~al.:
  An image is worth 16x16 words: Transformers for image recognition at scale.
  arXiv preprint arXiv:2010.11929  (2020)

\bibitem{TraftonDrew2013InformaticsIR}
Drew, T., Evans, K.K., V{\~o}, M.L.H., Jacobson, F.L., Wolfe, J.M.: Informatics
  in radiology: What can you see in a single glance and how might this guide
  visual search in medical images? Radiographics  \textbf{33},  263--274 (2013)

\bibitem{du2021towards}
Du, M., Manjunatha, V., Jain, R., Deshpande, R., Dernoncourt, F., Gu, J., Sun,
  T., Hu, X.: Towards interpreting and mitigating shortcut learning behavior of
  nlu models. arXiv preprint arXiv:2103.06922  (2021)

\bibitem{geirhos2020shortcut}
Geirhos, R., Jacobsen, J.H., Michaelis, C., Zemel, R., Brendel, W., Bethge, M.,
  Wichmann, F.A.: Shortcut learning in deep neural networks. Nature Machine
  Intelligence  \textbf{2}(11),  665--673 (2020)

\bibitem{han2022survey}
Han, K., Wang, Y., Chen, H., Chen, X., Guo, J., Liu, Z., Tang, Y., Xiao, A.,
  Xu, C., Xu, Y., et~al.: A survey on vision transformer. IEEE Transactions on
  Pattern Analysis and Machine Intelligence  (2022)

\bibitem{hassani2021escaping}
Hassani, A., Walton, S., Shah, N., Abuduweili, A., Li, J., Shi, H.: Escaping
  the big data paradigm with compact transformers. arXiv preprint
  arXiv:2104.05704  (2021)

\bibitem{he2021masked}
He, K., Chen, X., Xie, S., Li, Y., Dollár, P., Girshick, R.: Masked
  autoencoders are scalable vision learners (2021)

\bibitem{He_2016_CVPR}
He, K., Zhang, X., Ren, S., Sun, J.: Deep residual learning for image
  recognition. In: Proceedings of the IEEE Conference on Computer Vision and
  Pattern Recognition (CVPR) (June 2016)

\bibitem{2020Creation}
Karargyris, A., Kashyap, S., Lourentzou, I., Wu, J.T., Sharma, A., Tong, M.,
  Abedin, S., Beymer, D., Mukherjee, V., Krupinski, E.A., et~al.: Creation and
  validation of a chest x-ray dataset with eye-tracking and report dictation
  for ai development. Scientific data  \textbf{8}(1),  1--18 (2021)

\bibitem{AlexandrosKarargyris2021CreationAV}
Karargyris, A., Kashyap, S., Lourentzou, I., Wu, J.T., Sharma, A., Tong, M.H.,
  Abedin, S., Beymer, D., Mukherjee, V., Krupinski, E.A., Moradi, M.: Creation
  and validation of a chest x-ray dataset with eye-tracking and report
  dictation for ai development. Scientific Data  \textbf{8},  92--92 (2021)

\bibitem{EllenMKok2012LookingIT}
Kok, E.M., de~Bruin, A.B.H., Robben, S.G.F., van Merri{\"e}nboer, J.J.G.:
  Looking in the same manner but seeing it differently: Bottom‐up and
  expertise effects in radiology. Applied Cognitive Psychology  \textbf{26},
  854--862 (2012)

\bibitem{EllenMKok2017BeforeYV}
Kok, E.M., Jarodzka, H.: Before your very eyes: the value and limitations of
  eye tracking in medical education. Medical Education  \textbf{51},  114--122
  (2017)

\bibitem{krishnan2021vision}
Krishnan, K.S., Krishnan, K.S.: Vision transformer based covid-19 detection
  using chest x-rays. In: 2021 6th International Conference on Signal
  Processing, Computing and Control (ISPCC). pp. 644--648. IEEE (2021)

\bibitem{2010Current}
Krupinski, E.A.: Current perspectives in medical image perception. Attention
  Perception \& Psychophysics  \textbf{72}(5),  1205--1217 (2010)

\bibitem{ElizabethAKrupinski1996VisualSP}
Krupinski, E.A.: Visual scanning patterns of radiologists searching mammograms.
  Academic Radiology  \textbf{3},  137--144 (1996)

\bibitem{HaroldLKundel2007HolisticCO}
Kundel, H.L., Nodine, C.F., Conant, E.F., Weinstein, S.P.: Holistic component
  of image perception in mammogram interpretation: gaze-tracking study.
  Radiology  \textbf{242},  396--402 (2007)

\bibitem{lecun2015deep}
LeCun, Y., Bengio, Y., Hinton, G.: Deep learning. nature  \textbf{521}(7553),
  436--444 (2015)

\bibitem{li2021few}
Li, S., Sui, X., Fu, J., Fu, H., Luo, X., Feng, Y., Xu, X., Liu, Y., Ting,
  D.S., Goh, R.S.M.: Few-shot domain adaptation with polymorphic transformers.
  In: International Conference on Medical Image Computing and Computer-Assisted
  Intervention. pp. 330--340. Springer (2021)

\bibitem{li2021localvit}
Li, Y., Zhang, K., Cao, J., Timofte, R., Van~Gool, L.: Localvit: Bringing
  locality to vision transformers. arXiv preprint arXiv:2104.05707  (2021)

\bibitem{LauraLiberman2002BreastIR}
Liberman, L., Menell, J.H.: Breast imaging reporting and data system (bi-rads).
  Radiologic Clinics of North America  \textbf{40},  409--430 (2002)

\bibitem{liu2021swin}
Liu, Z., Lin, Y., Cao, Y., Hu, H., Wei, Y., Zhang, Z., Lin, S., Guo, B.: Swin
  transformer: Hierarchical vision transformer using shifted windows. In:
  Proceedings of the IEEE/CVF International Conference on Computer Vision. pp.
  10012--10022 (2021)

\bibitem{luo2021rethinking}
Luo, L., Chen, H., Xiao, Y., Zhou, Y., Wang, X., Vardhanabhuti, V., Wu, M.,
  Heng, P.A.: Rethinking annotation granularity for overcoming deep shortcut
  learning: A retrospective study on chest radiographs. arXiv preprint
  arXiv:2104.10553  (2021)

\bibitem{luo2021rectifying}
Luo, X., Wei, L., Wen, L., Yang, J., Xie, L., Xu, Z., Tian, Q.: Rectifying the
  shortcut learning of background for few-shot learning. Advances in Neural
  Information Processing Systems  \textbf{34} (2021)

\bibitem{ma2019affinitynet}
Ma, T., Zhang, A.: Affinitynet: semi-supervised few-shot learning for disease
  type prediction. In: Proceedings of the AAAI conference on artificial
  intelligence. vol.~33, pp. 1069--1076 (2019)

\bibitem{SuneetaMall2018ModelingVS}
Mall, S., Brennan, P.C., Mello-Thoms, C.: Modeling visual search behavior of
  breast radiologists using a deep convolution neural network. Journal of
  medical imaging  \textbf{5},  035502--035502 (2018)

\bibitem{SuneetaMall2019MissedCA}
Mall, S., Krupinski, E., Mello-Thoms, C.: Missed cancer and visual search of
  mammograms: what feature-based machine-learning can tell us that
  deep-convolution learning cannot. In: Medical Imaging 2019: Image Perception,
  Observer Performance, and Technology Assessment. vol. 10952, pp. 281--287.
  SPIE (2019)

\bibitem{mccoy2019right}
McCoy, R.T., Pavlick, E., Linzen, T.: Right for the wrong reasons: Diagnosing
  syntactic heuristics in natural language inference. arXiv preprint
  arXiv:1902.01007  (2019)

\bibitem{minderer2020automatic}
Minderer, M., Bachem, O., Houlsby, N., Tschannen, M.: Automatic shortcut
  removal for self-supervised representation learning. In: International
  Conference on Machine Learning. pp. 6927--6937. PMLR (2020)

\bibitem{mondal2021xvitcos}
Mondal, A.K., Bhattacharjee, A., Singla, P., Prathosh, A.: xvitcos: Explainable
  vision transformer based covid-19 screening using radiography. IEEE Journal
  of Translational Engineering in Health and Medicine  \textbf{10},  1--10
  (2021)

\bibitem{InsMoreira2012INbreastTA}
Moreira, I., Amaral, I., Domingues, I., Cardoso, A., Cardoso, M.J., Cardoso,
  J.S.: Inbreast: toward a full-field digital mammographic database. Academic
  Radiology  \textbf{19},  236--248 (2012)

\bibitem{nauta2022uncovering}
Nauta, M., Walsh, R., Dubowski, A., Seifert, C.: Uncovering and correcting
  shortcut learning in machine learning models for skin cancer diagnosis.
  Diagnostics  \textbf{12}(1), ~40 (2022)

\bibitem{niven2019probing}
Niven, T., Kao, H.Y.: Probing neural network comprehension of natural language
  arguments. arXiv preprint arXiv:1907.07355  (2019)

\bibitem{CalvinFNodine1987UsingEM}
Nodine, C.F., Kundel, H.L.: Using eye movements to study visual search and to
  improve tumor detection. Radiographics  \textbf{7},  1241--1250 (1987)

\bibitem{nystrom2010adaptive}
Nystr{\"o}m, M., Holmqvist, K.: An adaptive algorithm for fixation, saccade,
  and glissade detection in eyetracking data. Behavior research methods
  \textbf{42}(1),  188--204 (2010)

\bibitem{park2021vision}
Park, S., Kim, G., Oh, Y., Seo, J.B., Lee, S.M., Kim, J.H., Moon, S., Lim,
  J.K., Ye, J.C.: Vision transformer for covid-19 cxr diagnosis using chest
  x-ray feature corpus. arXiv preprint arXiv:2103.07055  (2021)

\bibitem{puch2019few}
Puch, S., S{\'a}nchez, I., Rowe, M.: Few-shot learning with deep triplet
  networks for brain imaging modality recognition. In: Domain Adaptation and
  Representation Transfer and Medical Image Learning with Less Labels and
  Imperfect Data, pp. 181--189. Springer (2019)

\bibitem{robinson2021deep}
Robinson, C., Trivedi, A., Blazes, M., Ortiz, A., Desbiens, J., Gupta, S.,
  Dodhia, R., Bhatraju, P.K., Liles, W.C., Lee, A., et~al.: Deep learning
  models for covid-19 chest x-ray classification: Preventing shortcut learning
  using feature disentanglement. medRxiv  (2021)

\bibitem{robinson1968oculomotor}
Robinson, D.A.: The oculomotor control system: A review. Proceedings of the
  IEEE  \textbf{56}(6),  1032--1049 (1968)

\bibitem{OlgaRussakovsky2014ImageNetLS}
Russakovsky, O., Deng, J., Su, H., Krause, J., Satheesh, S., Ma, S., Huang, Z.,
  Karpathy, A., Khosla, A., Bernstein, M.S., Berg, A.C., Fei-Fei, L.: Imagenet
  large scale visual recognition challenge. arXiv: Computer Vision and Pattern
  Recognition  (2014)

\bibitem{KhaledSaab2021ObservationalSF}
Saab, K., Hooper, S.M., Sohoni, N.S., Parmar, J., Pogatchnik, B.P., Wu, S.,
  Dunnmon, J., Zhang, H., Rubin, D.L., R{\'e}, C.: Observational supervision
  for medical image classification using gaze data. In: Medical Image Computing
  and Computer-Assisted Intervention (2021)

\bibitem{Selvaraju_2017_ICCV}
Selvaraju, R.R., Cogswell, M., Das, A., Vedantam, R., Parikh, D., Batra, D.:
  Grad-cam: Visual explanations from deep networks via gradient-based
  localization. In: Proceedings of the IEEE International Conference on
  Computer Vision (ICCV) (Oct 2017)

\bibitem{shamshad2022transformers}
Shamshad, F., Khan, S., Zamir, S.W., Khan, M.H., Hayat, M., Khan, F.S., Fu, H.:
  Transformers in medical imaging: A survey. arXiv preprint arXiv:2201.09873
  (2022)

\bibitem{shen2021towards}
Shen, X., Lam, W.: Towards domain-generalizable paraphrase identification by
  avoiding the shortcut learning. In: Proceedings of the International
  Conference on Recent Advances in Natural Language Processing (RANLP 2021).
  pp. 1318--1325 (2021)

\bibitem{shome2021covid}
Shome, D., Kar, T., Mohanty, S.N., Tiwari, P., Muhammad, K., AlTameem, A.,
  Zhang, Y., Saudagar, A.K.J.: Covid-transformer: Interpretable covid-19
  detection using vision transformer for healthcare. International Journal of
  Environmental Research and Public Health  \textbf{18}(21),  11086 (2021)

\bibitem{shorfuzzaman2021metacovid}
Shorfuzzaman, M., Hossain, M.S.: Metacovid: A siamese neural network framework
  with contrastive loss for n-shot diagnosis of covid-19 patients. Pattern
  recognition  \textbf{113},  107700 (2021)

\bibitem{SIIM-ACR}
{SIIM-ACR} pneumothorax segmentation (2020), [online] Available:
  \url{https://www.kaggle.com/c/siim-acr-pneumothorax-segmentation}

\bibitem{RichardGSwensson1980ATD}
Swensson, R.G.: A two-stage detection model applied to skilled visual search by
  radiologists. Attention Perception \& Psychophysics  \textbf{27},  11--16
  (1980)

\bibitem{touvron2021training}
Touvron, H., Cord, M., Douze, M., Massa, F., Sablayrolles, A., J{\'e}gou, H.:
  Training data-efficient image transformers \& distillation through attention.
  In: International Conference on Machine Learning. pp. 10347--10357. PMLR
  (2021)

\bibitem{wang2022follow}
Wang, S., Ouyang, X., Liu, T., Wang, Q., Shen, D.: Follow my eye: Using gaze to
  supervise computer-aided diagnosis. IEEE Transactions on Medical Imaging
  (2022)

\bibitem{wang2020generalizing}
Wang, Y., Yao, Q., Kwok, J.T., Ni, L.M.: Generalizing from a few examples: A
  survey on few-shot learning. ACM computing surveys (csur)  \textbf{53}(3),
  1--34 (2020)

\bibitem{xiao2020noise}
Xiao, K., Engstrom, L., Ilyas, A., Madry, A.: Noise or signal: The role of
  image backgrounds in object recognition. arXiv preprint arXiv:2006.09994
  (2020)

\bibitem{yuan2021incorporating}
Yuan, K., Guo, S., Liu, Z., Zhou, A., Yu, F., Wu, W.: Incorporating convolution
  designs into visual transformers. In: Proceedings of the IEEE/CVF
  International Conference on Computer Vision. pp. 579--588 (2021)

\bibitem{zech2018variable}
Zech, J.R., Badgeley, M.A., Liu, M., Costa, A.B., Titano, J.J., Oermann, E.K.:
  Variable generalization performance of a deep learning model to detect
  pneumonia in chest radiographs: a cross-sectional study. PLoS medicine
  \textbf{15}(11),  e1002683 (2018)

\bibitem{zhang2021aggregating}
Zhang, Z., Zhang, H., Zhao, L., Chen, T., Pfister, T.: Aggregating nested
  transformers. arXiv preprint arXiv:2105.12723  (2021)

\bibitem{CLAHE}
Zuiderveld, K.: Contrast limited adaptive histograph equalization. Graphic Gems
  IV p. 474–485 (1994)

\end{thebibliography}
\end{document}